\documentclass[runningheads]{llncs}
\usepackage[T1]{fontenc}
\usepackage{graphicx,verbatim}
\usepackage{graphicx}
\usepackage{amsmath,amssymb} 
\usepackage{booktabs}
\usepackage{enumitem}
\usepackage{algorithm}
\usepackage{algpseudocode}
\usepackage{pifont}
\usepackage[table]{xcolor}
\usepackage[colorlinks,urlcolor=blue,citecolor=blue]{hyperref}
\usepackage{multirow}
\usepackage{subcaption}
\usepackage{url}
\usepackage{mathtools}
\usepackage{bm}
\usepackage[nameinlink]{cleveref}

\newcommand{\x}{\bm{\mathrm{x}}}

\newcommand{\xy}{\bm{\mathrm{x}}_{\tiny \text{Y}}}
\newcommand{\xcb}{\bm{\mathrm{x}}_{\tiny \text{Cb}}}
\newcommand{\xcr}{\bm{\mathrm{x}}_{\tiny \text{Cr}}}
\newcommand{\Sy}{{\mathrm{X}}}

\usepackage{orcidlink}
\renewcommand{\orcidID}[1]{\orcidlink{#1}}
\newcommand\mapsfrom{\mathrel{\reflectbox{\ensuremath{\mapsto}}}}

\definecolor{codeblue}{rgb}{0.25, 0.5, 0.5}
\definecolor{OurResult}{RGB}{203, 104, 67}

\usepackage{marvosym}

\begin{document}
\title{PhaseAT: Fourier Phase Adversarial Training for Medical Image Domain Generalization}

\author{Ahmed Sharshar \orcidID{0000-0003-2280-5240}(\Letter) \and
Asif Hanif \orcidID{0000-0002-6705-149X} \and
Naveen Kumar Kummari \orcidID{0000-0003-4250-4429} \and
Mohammad Yaqub \orcidID{0000-0001-6896-1105} \and
Mohsen Guizani \orcidID{0000-0002-8972-8094}}
\authorrunning{Sharshar et al.}
\institute{Division of Computing and Mathematical Sciences \\
Mohamed bin Zayed University of Artificial Intelligence, Abu Dhabi, UAE\\
\email{\{ahmed.sharshar, asif.hanif, naveen.kummari, mohammad.yaqub, mohsen.guizani\}@mbzuai.ac.ae}}

\maketitle              
\begin{abstract}
Reliable clinical deployment of deep medical image models is hindered by distribution shifts across scanners, sites, and acquisition protocols. Existing domain generalization (DG) methods often focus on style or intensity diversification, but they can still leave networks dependent on domain-specific texture correlations. Inspired by evidence that Fourier phase encodes semantic structure, we introduce \textsc{PhaseAT}, a phase-aware adversarial training framework for medical DG. \textsc{PhaseAT} forms phase-perturbed training views in the Fourier domain by iteratively updating a bounded phase perturbation while keeping the amplitude spectrum unchanged, thereby stressing spatial organization under matched appearance statistics. Perturbations are applied only to the luminance channel in YCbCr color space to avoid chromatic artifacts. Additionally, a simple phase-saliency mask concentrates updates on the most influential frequencies. The model is trained with a weighted combination of losses on clean and phase-perturbed samples, supporting both single-source and multi-source DG. We validate our method on two challenging medical datasets and demonstrate that \textsc{PhaseAT} achieves over 20\% improvement in single-source domain generalization, outperforming several state-of-the-art DG methods. The code implementation is available at: \url{https://github.com/ahmed-sharshar/PhaseAT}.

\keywords{Domain Generalization  \and Adversarial Training \and \\ Phase-Aware Perturbation.}

\end{abstract}

\section{Introduction}
\label{sec:intro}

Deep learning has become the dominant paradigm for medical image analysis, yet standard training pipelines still rely on an \emph{i.i.d.} assumption that rarely holds in clinical practice. In reality, medical imaging systems routinely experience distribution shifts induced by scanner vendors, acquisition protocols, reconstruction pipelines, and diverse patient populations~\cite{Safdari2025MixStyleFlow}. From a spectral viewpoint, such shifts are often reflected in changes to the Fourier \emph{amplitude} spectrum, which captures low-level appearance statistics (e.g., intensity and texture), while the Fourier phase largely preserves spatial organization and geometry~\cite{Lee2023DACNorm}. Since standard CNNs are biased toward texture cues~\cite{geirhos2019imagenettexture}, they frequently overfit to these unstable amplitude-driven correlations, causing brittle performance when deployed outside the source environment.

\emph{Domain Generalization} (DG) targets this challenge by learning predictors that remain accurate on unseen target domains without requiring target data during training. This distinction is crucial in medicine, where privacy constraints and logistical silos limit multi-institutional data aggregation, often leaving practitioners with only a small number of source environments~\cite{Su2023SSDA}. Multi-center benchmarks further underscore that such shifts are routine rather than exceptional in medical imaging, with substantial gaps under cross-hospital evaluation~\cite{bandi2019camelyon17}. The central DG question remains: how can we learn features that stay stable when imaging style changes but clinical semantics do not?

Prior medical DG methods have made steady progress, though most target only subsets of shift factors. Existing approaches broadly fall into three families. First, \emph{style/intensity diversification} methods simulate scanner-driven appearance variation, for example via adversarial style composition~\cite{Chen2022MaxStyle}, content--style augmentation~\cite{CheXi_ConStyX_MICCAI2025}, or monotonic adversarial intensity attacks~\cite{Zhang2026AdverIN}; others use representation-driven objectives to regularize style/content separation~\cite{siddiquee2025condisr}. Second, \emph{frequency-domain} approaches exploit spectral structure through Fourier mixing, amplitude synthesis, or reconstruction constraints to reduce scanner-specific artifacts~\cite{Qiao_2024_MICCAI}. Third, \emph{structure-consistent} methods that focus on anatomy-aligned cues, such as via anatomy-consistent pseudo-modalities~\cite{Hu2024MAP} or edge-guided generalization~\cite{Jiang2024EdgeDG}. Despite these advances, performance can remain sensitive to hyperparameter choices, and carefully tuned baselines are often surprisingly strong~\cite{Teterwak_2025_ICLR}. This motivates mechanisms that more directly reflect how medical shifts arise in practice~\cite{Zhang_2025_ICCV}, especially in limited-source regimes where the inductive bias of the training objective becomes paramount~\cite{Yi_2024_MICCAI}.

A promising route to robustness is \emph{Adversarial Training} (AT), but the perturbation model must reflect clinically relevant shifts. Standard pixel-space perturbations tend to emphasize high-frequency noise, potentially reinforcing a model's texture bias~\cite{geirhos2019imagenettexture}. A Fourier viewpoint enables more targeted control: the \emph{amplitude} spectrum is closely tied to low-level appearance statistics, while \emph{phase} governs spatial organization~\cite{Lee2023DACNorm}. Indeed, phase is the primary carrier of recognizable structure~\cite{zhou2023phaseaware}. Recent work further shows that improving stability to phase manipulations can enhance structural robustness~\cite{zhou2023phaseaware}, and frequency-constrained perturbations can effectively expose model vulnerabilities in medical tasks~\cite{zhang2024fourier}. Together, these observations suggest that a \emph{phase-focused} robustness objective can better match clinical distribution shifts while discouraging shortcut learning.

\begin{figure*}[t]
    \centering
    \includegraphics[width=0.98\textwidth]{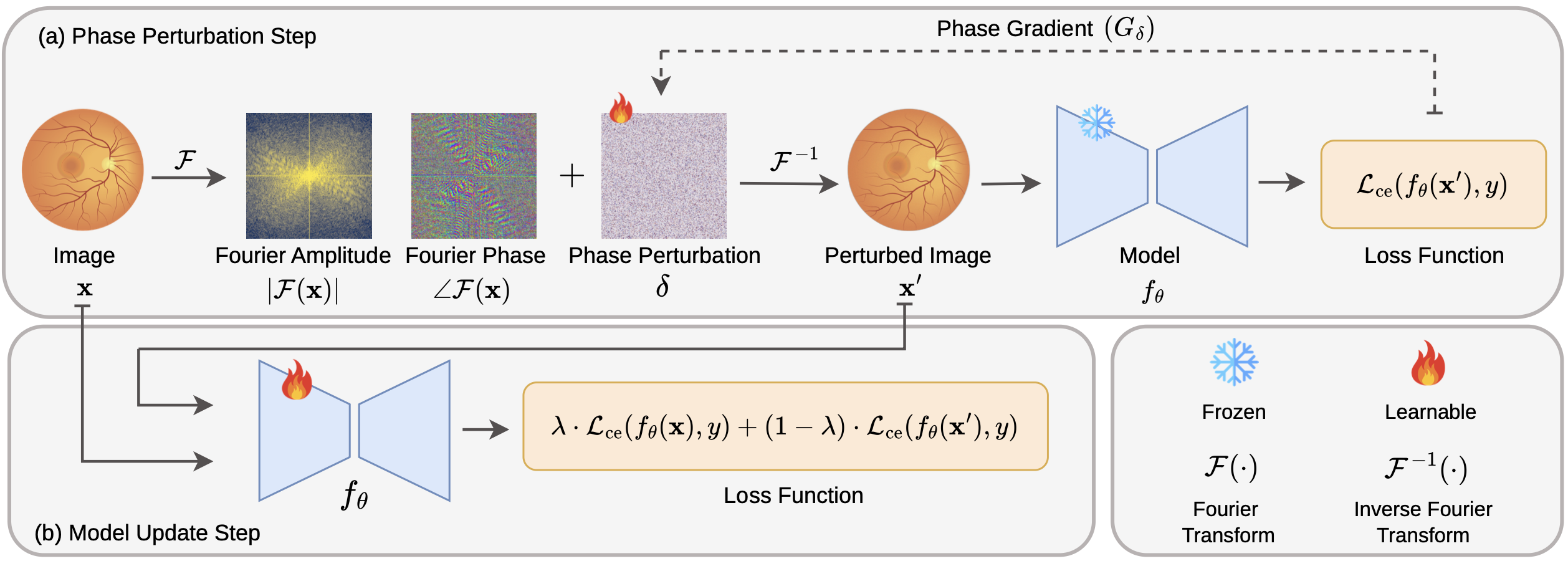} 
    \caption{\textsc{PhaseAT Pipeline}
    \textbf{(a)} \textit{Phase Perturbation Step (inner maximization):} Adversarial phase perturbations are applied in the Fourier domain by updating the phase component of the luminance channel, generating phase-perturbed view $\mathrm{x}'$.
    \textbf{(b)} \textit{Model Update Step (outer minimization):} Model parameters are optimized using a joint objective that incorporates both clean and phase-perturbed samples to improve robustness.}
    \label{fig:main_fig}
\end{figure*}

In this work, we propose \emph{Phase-Aware Adversarial Training} (\textsc{PhaseAT}) for medical domain generalization, Figure \ref{fig:main_fig}. \textsc{PhaseAT} synthesizes training views by adversarially perturbing the Fourier \emph{phase}, thereby challenging anatomy-relevant spatial organization, while \emph{fixing} the Fourier amplitude to preserve per-frequency energy statistics (appearance). To avoid unnecessary chromatic artifacts and target geometry directly, we apply perturbations only to the luminance channel in YCbCr. The resulting phase-perturbed samples serve as hard training instances that complement recent appearance diversification and Fourier-based DG strategies~\cite{Zhang2026AdverIN,Qiao_2024_MICCAI} by explicitly encouraging invariance to geometry-focused spectral variations under comparable appearance statistics. Unlike Fourier-based DG that varies the \emph{amplitude} spectrum~\cite{Qiao_2024_MICCAI} or pixel-space augmentation, \textsc{PhaseAT} fixes amplitude and perturbs only \emph{phase}, isolating geometry from appearance.

\noindent\textbf{Contributions.} We propose \textsc{PhaseAT}, a phase-aware adversarial training for medical image \emph{single-source} and \emph{multi-source} domain generalization, using robustness as a mechanism to improve out-of-domain performance. We introduce a \emph{phase-constrained} spectral perturbation scheme that perturbs Fourier phase and applies perturbations in luminance to reduce artifacts under domain shift. Finally, we provide a comprehensive evaluation across different datasets, demonstrating consistent improvements over a range of recent DG baselines.

\section{Methodology}

To instantiate the phase-aware robustness objective described in Section~\ref{sec:intro}, we generate phase-perturbed samples in the Fourier domain while preserving the amplitude spectrum, and incorporate these samples during training to enhance domain generalization. Specifically, spectral perturbations are applied only to the luminance channel in the YCbCr color space. This design choice is motivated by three considerations. First, YCbCr allocates higher effective resolution to luminance than to chrominance, reflecting the fact that fine spatial details are primarily encoded in the $(\mathrm{Y})$ channel. Second, human visual sensitivity to high-frequency spatial structure is substantially greater for achromatic (luminance) variations than for chromatic ones, meaning that edges and geometric information are predominantly conveyed by luminance~\cite{pestana2020ychannel}. Third, prior studies have shown that gradient-based perturbations tend to concentrate in the $(\mathrm{Y})$ channel when images are represented in YCbCr, suggesting that restricting the inner maximization to luminance is both effective and avoids unnecessary chromatic distortions~\cite{pestana2020ychannel}. From an optimization perspective, operating on a single channel ($\mathrm{Y}$) rather than three (RGB) reduces the dimensionality of the inner maximization problem. This not only improves computational efficiency but also produces more structured perturbations with fewer color artifacts.

\subsection{Phase-Constrained Spectral Perturbations}

Consider an RGB image $\x \in [0,1]^{H \times W \times 3}$ and its corresponding label $y$, with pixel values normalized to $[0,1]$ for each channel. The RGB image $\x$ is first converted into the YCbCr color space, producing $(\xy,\xcb,\xcr) \in [0,1]^{H \times W \times 3}$, where the $\xy$ is luminance channel. The luminance channel is transformed into the Fourier domain, yielding the complex spectrum $\Sy = \mathcal{F}(\xy) \in \mathbb{C}^{H \times W}$, which can be decomposed into amplitude and phase components as $\Sy = |\Sy| e^{(j\angle\Sy)}$. To generate a phase-perturbed representation, we introduce a perturbation $\delta \in \mathbb{R}^{H \times W}$ and define the modified spectrum as: $
\Sy^{\prime} = |\Sy| e^{j(\angle \Sy + \delta)}$, thereby preserving the original amplitude spectrum $|\Sy|$ while altering only the phase component. Applying the inverse Fourier transform yields the perturbed luminance channel, $\xy^{\prime} = \mathcal{F}^{-1}(\Sy^{\prime})$. The perturbed luminance channel is combined with the original chrominance channels to form  $(\xy^{\prime}; \xcb; \xcr)$, which is then transformed back to RGB space, yielding the perturbed image $\x^{\prime}$. To obtain adversarial phase perturbations, $\delta$ is optimized by maximizing the cross-entropy loss:
\begin{equation}
\underset{\delta}{\operatorname{maximize}}
\;\;
\mathcal{L}_{\mathrm{ce}}\!\left(f_\theta(\x^{\prime}), y\right)
\quad
\text{s.t.}
\quad
\|\delta\|_{\infty} \le \epsilon,
\end{equation}
\noindent where $\epsilon$ specifies the maximum admissible phase deviation. Since $\Sy$ is complex-valued, the phase update is computed directly in the Fourier domain. Specifically, we derive the gradient of the loss with respect to the spectrum phase and use the resulting phase-direction gradient as a proxy signal for updating $\delta$. Using Wirtinger calculus~\cite{wirtinger}, phase-direction proxy gradient ${G}_{\delta}$ can be expressed as:
\begin{equation}
\label{eq:grad_phi}
{G}_{\delta} \triangleq -2 \cdot \Im(G_{\Sy} \odot \Sy),
\end{equation}
where $G_{\Sy}$ denotes the gradient of the loss w.r.t the complex spectrum $\Sy$, $\odot$ is the Hadamard product and $\Im(\cdot)$ denotes the imaginary part. To ensure a stable and well-defined phase update, we compute a phase-saliency soft mask $W_{\delta} \in [0,1]^{H \times W}$ by applying min--max normalization to the magnitude of the phase-direction gradient, $|{G}_{\delta}|$. This mask captures the loss sensitivity to phase perturbations, upweighting frequency components that induce larger loss changes. It thus focuses the phase update on structurally informative, adversarially relevant frequencies. Phase perturbation is updated as follows:
\begin{equation}
\label{eq:update_phi}
\delta \;\mapsfrom\; 
\mathcal{P}_{[-\epsilon,\epsilon]}\!\Big(
\delta + \eta \cdot 
W_{\delta} \odot \operatorname{sign}({G}_{\delta})
\Big),
\end{equation}
where $\eta$ denotes the update step size, $W_{\delta} \in [0,1]^{H \times W}$ is the phase-saliency weighting mask, $\mathcal{P}_{[-\epsilon,\epsilon]}(\cdot)$ denotes element-wise projection onto the $\ell_{\infty}$ ball of radius $\epsilon$, and $\operatorname{sign}(\cdot)$ is the element-wise sign operator.

\subsection{Phase-Aware Adversarial Training}
\label{sec:phaseat}

Given a clean sample $\x$ and its phase-perturbed counterpart $\x'$, we update model parameters by minimizing a convex combination of clean and perturbed losses:
\begin{equation}
\label{eq:phaseat_loss}
\underset{\theta}{\operatorname{minimize}}
\;\;\;  \lambda\cdot\mathcal{L}_{\mathrm{ce}}(f_\theta(\x),y) + (1-\lambda)\cdot\mathcal{L}_{\mathrm{ce}}(f_\theta(\x'),y),  
\end{equation}

where $\lambda\in[0,1]$ controls the trade-off between standard and robust accuracy. Algorithm~\ref{alg:phaseat_train} summarizes the full training procedure; the same routine applies to single-source and multi-source DG by defining $\mathcal{D}_s$ as the available source data.

\begin{algorithm}[t]
\caption{\textsc{PhaseAT}: Phase-Adv. Training for Domain Generalization}
\label{alg:phaseat_train}
\begin{algorithmic}[1]
\State Source Data $\mathcal{D}_s$, Model $f_\theta$, Epochs $N$, Steps $K$, Step Size $\eta$, Budget $\epsilon$, Loss Weight $\lambda$, Learning Rate $\gamma$, FFT $\mathcal{F}$, IFFT $\mathcal{F}^{-1}$, Projection $\mathcal{P}_{[-\epsilon,\epsilon]}$, Imaginary Part $\Im$
\Statex
\For{$\mathrm{epoch}=1,\ldots,N$}
    \State $(\x,y)\sim \mathcal{D}_s$ \Comment{Sample a mini-batch (single-source or pooled multi-source)}
    \State $(\xy,\xcb,\xcr) \gets \texttt{RGB2YCbCr}(\x)$ \Comment{Convert from RGB to YCbCr color space}
    \State $\Sy \gets \mathcal{F}(\xy) = |\Sy|e^{j\angle\Sy}$ \Comment{Luminance channel spectrum}
    \State $\delta \gets \mathbf{0}\in\mathbb{R}^{H\times W}$ \Comment{Initialize phase perturbation}
    \For{$k=1,\ldots,K$}
        \State $\Sy' \mapsfrom |\Sy|e^{j(\angle\Sy+\delta)}$ \Comment{Add perturbation to phase}
        \State $\mathcal{F}^{-1}(\Sy') \to \xy' \to \texttt{YCbCr2RGB}([\xy',\xcb,\xcr]) \to \x'$ \Comment{Fourier-to-Spatial}
        \State Compute cross-entropy loss:  $\mathcal{L}_{\mathrm{ce}}(f_\theta(\x'),y)$
        \State $G_{\delta} \triangleq -2\cdot \Im(\nabla_{\Sy}\odot \Sy)$ \Comment{Phase-direction gradient (Eq.~\ref{eq:grad_phi})}
        \State $\delta \mapsfrom \mathcal{P}_{[-\epsilon,\epsilon]}\!\big(\delta + \eta\cdot W_{\delta} \odot \mathrm{sign}(G_{\delta})\big)$ \Comment{Update $\delta$ with gradient ascent (Eq.~\ref{eq:update_phi})}
    \EndFor
    \State $\ell \mapsfrom \lambda\,\mathcal{L}_{\mathrm{ce}}(f_\theta(\x),y) + (1-\lambda)\,\mathcal{L}_{\mathrm{ce}}(f_\theta(\x'),y)$ \Comment{Training loss (Eq.~\ref{eq:phaseat_loss})}
    \State $\theta \mapsfrom \theta - \gamma \nabla_\theta \ell$ \Comment{Update model parameters via gradient descent}
\EndFor
\Statex
\State \textbf{Return:} Adversarially trained robust model $f_\theta$
\end{algorithmic}
\end{algorithm}

\section{Experimental Setup}
\label{sec:exp}

\noindent\textbf{Datasets.}
We evaluate on two medical benchmarks with strong cross-domain variation. \textit{Camelyon17-WILDS} ~\cite{bandi2019camelyon17} contains histopathology data from \textit{five medical centers} (domains). 
\textit{Diabetic Retinopathy (DR)} combines Aptos~\cite{aptos2019blindness}, EyePACS~\cite{kaggle2015dr}, Messidor~\cite{decenciere2014messidor}, and Messidor-2~\cite{messidor2} as distinct domains for multi-class severity prediction.

\noindent\textbf{Evaluation Protocol.}
We follow the \textit{DomainBed} methodology~\cite{gulrajani2020domainbed} which uses training-domain validation for model selection. We report accuracy on unseen domains in two regimes: (1) Single-domain generalization (SDG), where we train on one source and evaluate on the rest, and (2) Multi-source generalization (MSDG), using a leave-one-domain-out protocol.

\noindent\textbf{Implementation Details.}
\textit{Backbones:} We primarily use DenseNet-121 initialized from scratch. For architectural ablation and stability analysis, we also report results using ResNet-18, ViT-Base, and ImageNet-pretrained DenseNet-121, where specified.
\textit{Training:} Models are trained for 30 epochs on a single NVIDIA A6000 GPU, which is sufficient for convergence under our schedule (validation accuracy typically plateaus by the final full-attack phase). We use a 5-epoch clean warm-up, followed by a linear ramp-up of the adversarial strength/budget until epoch 25, then 5 epochs of full attack. We use K=5 inner maximization steps (Algorithm 1), Weighting ($\lambda$) = 0.5, and a step size of 0.2, and study their sensitivity. Results are averaged over 5 fixed random seeds.

\begin{table*}[t]
\centering
\caption{Performance comparison in the Single Domain Generalization (SDG) setting. Values are the accuracy(\%) mean $\pm$ standard deviation. Best and second-best per column are in \textbf{bold} and \underline{underline}; $^{\dagger}$ marks a significant gain over the strongest baseline (paired $t$-test, $p{<}0.05$).}
\label{tab:full_comparison}
\footnotesize
\setlength{\tabcolsep}{2pt} 

\begin{minipage}[t]{0.52\textwidth}
\centering
\subcaption{Camelyon17-WILDS}
\resizebox{\textwidth}{!}{%
\begin{tabular}{lcccccc}
\toprule
\rowcolor{gray!10} \textbf{\textsc{Method}} & $\bm{C_0}$ & $\bm{C_1}$ & $\bm{C_2}$ & $\bm{C_3}$ & $\bm{C_4}$ & \textbf{\textsc{AVG.}} \\
\midrule
\multicolumn{7}{c}{\cellcolor{gray!5}\textit{Source Domain: $C_0$}} \\

ERM & \textbf{--} & $67.29_{\pm 1.2}$ & $65.89_{\pm 2.1}$ & $67.73_{\pm 1.5}$ & $56.50_{\pm 2.8}$ & $64.35_{\pm 1.9}$ \\
AdverIN~\cite{Zhang2026AdverIN} & \textbf{--} & $70.17_{\pm 1.4}$ & $69.30_{\pm 2.3}$ & $69.92_{\pm 1.7}$ & $60.26_{\pm 2.5}$ & $67.41_{\pm 2.0}$ \\
ConDiSR~\cite{siddiquee2025condisr} & \textbf{--} & $69.05_{\pm 1.1}$ & $67.45_{\pm 2.0}$ & $70.29_{\pm 1.6}$ & $\underline{60.48}_{\pm 2.4}$ & $66.82_{\pm 1.8}$ \\
RAS$^4$DG~\cite{Qiao_2024_MICCAI} & \textbf{--} & $\underline{71.15}_{\pm 1.3}$ & $\underline{70.34}_{\pm 2.2}$ & $\mathbf{76.28}_{\pm \mathbf{1.8}}$ & $58.46_{\pm 2.6}$ & $\underline{69.06}_{\pm 1.9}$ \\
\rowcolor{OurResult!15} \textbf{Ours} & \textbf{--} & $\mathbf{74.56}_{\pm \mathbf{1.0}}$ & $\mathbf{73.01}_{\pm \mathbf{1.5}}$ & $\underline{75.05}_{\pm 1.2}$ & $\mathbf{62.58}_{\pm \mathbf{1.8}}$ & $\mathbf{71.30}^{\dagger}_{\pm \mathbf{1.3}}$ \\
\midrule
\multicolumn{7}{c}{\cellcolor{gray!5}\textit{Source Domain: $C_1$}} \\

ERM & $61.87_{\pm 2.1}$ & \textbf{--} & $59.45_{\pm 2.5}$ & $63.01_{\pm 1.9}$ & $50.67_{\pm 2.7}$ & $58.75_{\pm 2.3}$ \\
AdverIN~\cite{Zhang2026AdverIN} & $65.42_{\pm 1.8}$ & \textbf{--} & $61.72_{\pm 2.4}$ & $65.94_{\pm 1.6}$ & $53.85_{\pm 2.5}$ & $61.73_{\pm 2.1}$ \\
ConDiSR~\cite{siddiquee2025condisr} & $64.54_{\pm 1.9}$ & \textbf{--} & $60.70_{\pm 2.3}$ & $64.83_{\pm 1.7}$ & $61.00_{\pm 2.2}$ & $62.77_{\pm 2.0}$ \\
RAS$^4$DG~\cite{Qiao_2024_MICCAI} & $\underline{73.39}_{\pm 1.5}$ & \textbf{--} & $\underline{73.05}_{\pm 1.8}$ & $\underline{76.48}_{\pm 1.4}$ & $\mathbf{65.49}_{\pm \mathbf{2.0}}$ & $\underline{72.10}_{\pm 1.7}$ \\
\rowcolor{OurResult!15} \textbf{Ours} & $\mathbf{78.26}_{\pm \mathbf{1.2}}$ & \textbf{--} & $\mathbf{75.20}_{\pm \mathbf{1.4}}$ & $\mathbf{79.70}_{\pm \mathbf{1.1}}$ & $\underline{64.08}_{\pm 1.6}$ & $\mathbf{74.31}^{\dagger}_{\pm \mathbf{1.3}}$ \\
\midrule
\multicolumn{7}{c}{\cellcolor{gray!5}\textit{Source Domain: $C_2$}} \\

ERM & $55.12_{\pm 2.6}$ & $47.46_{\pm 2.9}$ & \textbf{--} & $51.09_{\pm 2.4}$ & $54.65_{\pm 2.2}$ & $52.08_{\pm 2.5}$ \\
AdverIN~\cite{Zhang2026AdverIN} & $57.76_{\pm 2.3}$ & $51.36_{\pm 2.7}$ & \textbf{--} & $53.20_{\pm 2.1}$ & $58.02_{\pm 2.0}$ & $55.09_{\pm 2.3}$ \\
ConDiSR~\cite{siddiquee2025condisr} & $56.10_{\pm 2.4}$ & $51.91_{\pm 2.6}$ & \textbf{--} & $52.94_{\pm 2.2}$ & $57.34_{\pm 2.1}$ & $54.57_{\pm 2.4}$ \\
RAS$^4$DG~\cite{Qiao_2024_MICCAI} & $\underline{75.71}_{\pm 1.6}$ & $\mathbf{69.40}_{\pm \mathbf{1.9}}$ & \textbf{--} & $\underline{68.35}_{\pm 1.7}$ & $\underline{74.86}_{\pm 1.5}$ & $\underline{72.08}_{\pm 1.7}$ \\
\rowcolor{OurResult!15} \textbf{Ours} & $\mathbf{78.65}_{\pm \mathbf{1.3}}$ & $\underline{67.72}_{\pm 1.8}$ & \textbf{--} & $\mathbf{72.90}_{\pm \mathbf{1.4}}$ & $\mathbf{77.97}_{\pm \mathbf{1.2}}$ & $\mathbf{74.31}^{\dagger}_{\pm \mathbf{1.4}}$ \\
\midrule
\multicolumn{7}{c}{\cellcolor{gray!5}\textit{Source Domain: $C_3$}} \\

ERM & $53.49_{\pm 2.5}$ & $46.45_{\pm 2.8}$ & $54.41_{\pm 2.3}$ & \textbf{--} & $38.25_{\pm 2.9}$ & $48.15_{\pm 2.6}$ \\
AdverIN~\cite{Zhang2026AdverIN} & $55.54_{\pm 2.2}$ & $50.08_{\pm 2.5}$ & $57.19_{\pm 2.0}$ & \textbf{--} & $41.47_{\pm 2.7}$ & $51.07_{\pm 2.4}$ \\
ConDiSR~\cite{siddiquee2025condisr} & $57.11_{\pm 2.1}$ & $49.09_{\pm 2.6}$ & $56.44_{\pm 2.1}$ & \textbf{--} & $\underline{45.63}_{\pm 2.5}$ & $52.07_{\pm 2.3}$ \\
RAS$^4$DG~\cite{Qiao_2024_MICCAI} & $\underline{64.40}_{\pm 1.8}$ & $\underline{53.76}_{\pm 2.4}$ & $\mathbf{68.73}_{\pm \mathbf{1.7}}$ & \textbf{--} & $43.98_{\pm 2.6}$ & $\underline{57.72}_{\pm 2.1}$ \\
\rowcolor{OurResult!15} \textbf{Ours} & $\mathbf{66.63}_{\pm \mathbf{1.5}}$ & $\mathbf{57.86}_{\pm \mathbf{1.9}}$ & $\underline{67.78}_{\pm 1.4}$ & \textbf{--} & $\mathbf{47.65}_{\pm \mathbf{2.2}}$ & $\mathbf{59.98}_{\pm \mathbf{1.8}}$ \\
\midrule
\multicolumn{7}{c}{\cellcolor{gray!5}\textit{Source Domain: $C_4$}} \\

ERM & $56.65_{\pm 2.4}$ & $53.12_{\pm 2.5}$ & $55.26_{\pm 2.3}$ & $57.53_{\pm 2.2}$ & \textbf{--} & $55.64_{\pm 2.4}$ \\
AdverIN~\cite{Zhang2026AdverIN} & $\underline{59.56}_{\pm 2.1}$ & $56.18_{\pm 2.2}$ & $\underline{57.73}_{\pm 2.0}$ & $\mathbf{61.42}_{\pm \mathbf{1.9}}$ & \textbf{--} & $\mathbf{58.72}_{\pm \mathbf{2.0}}$ \\
ConDiSR~\cite{siddiquee2025condisr} & $59.35_{\pm 2.2}$ & $\underline{56.74}_{\pm 2.3}$ & $56.76_{\pm 2.1}$ & $\underline{60.98}_{\pm 1.9}$ & \textbf{--} & $58.46_{\pm 2.1}$ \\
RAS$^4$DG~\cite{Qiao_2024_MICCAI} & $56.18_{\pm 2.3}$ & $\mathbf{57.10}_{\pm \mathbf{2.1}}$ & $56.17_{\pm 2.2}$ & $56.28_{\pm 2.3}$ & \textbf{--} & $56.43_{\pm 2.2}$ \\
\rowcolor{OurResult!15} \textbf{Ours} & $\mathbf{59.70}_{\pm \mathbf{1.7}}$ & $55.98_{\pm 1.9}$ & $\mathbf{58.24}_{\pm \mathbf{1.8}}$ & $60.64_{\pm 1.6}$ & \textbf{--} & $\underline{58.64}_{\pm 1.8}$ \\
\bottomrule
\end{tabular}%
}
\end{minipage}
\hfill
\begin{minipage}[t]{0.46\textwidth}
\centering
\subcaption{Diabetic Retinopathy}
\resizebox{\textwidth}{!}{%
\begin{tabular}{lccccc}
\toprule
\rowcolor{gray!10} \textbf{\textsc{Method}} & \textbf{\textsc{Aptos}} & \textbf{\textsc{EyeP}} & \textbf{\textsc{Mess}} & \textbf{\textsc{Mess2}} & \textbf{\textsc{AVG.}} \\
\midrule
\multicolumn{6}{c}{\cellcolor{gray!5}\textit{Source: Aptos}} \\

ERM & \textbf{--} & $42.69_{\pm 2.5}$ & $30.32_{\pm 2.9}$ & $38.11_{\pm 2.6}$ & $37.04_{\pm 2.7}$ \\
AdverIN~\cite{Zhang2026AdverIN} & \textbf{--} & $45.03_{\pm 2.3}$ & $31.98_{\pm 2.8}$ & $40.20_{\pm 2.4}$ & $39.07_{\pm 2.5}$ \\
ConDiSR~\cite{siddiquee2025condisr} & \textbf{--} & $45.26_{\pm 2.2}$ & $32.14_{\pm 2.7}$ & $40.40_{\pm 2.3}$ & $39.27_{\pm 2.4}$ \\
RAS$^4$DG~\cite{Qiao_2024_MICCAI} & \textbf{--} & $\underline{48.14}_{\pm 2.0}$ & $\underline{33.60}_{\pm 2.6}$ & $\underline{42.75}_{\pm 2.2}$ & $\underline{41.50}_{\pm 2.3}$ \\
\rowcolor{OurResult!15} \textbf{Ours} & \textbf{--} & $\mathbf{51.85}_{\pm \mathbf{1.8}}$ & $\mathbf{36.82}_{\pm \mathbf{2.1}}$ & $\mathbf{46.28}_{\pm \mathbf{1.9}}$ & $\mathbf{44.99}^{\dagger}_{\pm \mathbf{1.9}}$ \\
\midrule
\multicolumn{6}{c}{\cellcolor{gray!5}\textit{Source: EyePACS}} \\

ERM & $39.42_{\pm 2.7}$ & \textbf{--} & $32.48_{\pm 2.8}$ & $39.48_{\pm 2.5}$ & $37.13_{\pm 2.7}$ \\
AdverIN~\cite{Zhang2026AdverIN} & $41.58_{\pm 2.5}$ & \textbf{--} & $34.26_{\pm 2.6}$ & $41.65_{\pm 2.3}$ & $39.16_{\pm 2.5}$ \\
ConDiSR~\cite{siddiquee2025condisr} & $41.79_{\pm 2.4}$ & \textbf{--} & $34.43_{\pm 2.7}$ & $41.86_{\pm 2.4}$ & $39.36_{\pm 2.5}$ \\
RAS$^4$DG~\cite{Qiao_2024_MICCAI} & $\underline{44.29}_{\pm 2.2}$ & \textbf{--} & $\underline{36.14}_{\pm 2.5}$ & $\underline{44.37}_{\pm 2.1}$ & $\underline{41.60}_{\pm 2.3}$ \\
\rowcolor{OurResult!15} \textbf{Ours} & $\mathbf{47.87}_{\pm \mathbf{1.9}}$ & \textbf{--} & $\mathbf{39.44}_{\pm \mathbf{2.2}}$ & $\mathbf{47.95}_{\pm \mathbf{1.8}}$ & $\mathbf{45.09}^{\dagger}_{\pm \mathbf{2.0}}$ \\
\midrule
\multicolumn{6}{c}{\cellcolor{gray!5}\textit{Source: Messidor}} \\

ERM & $41.91_{\pm 2.4}$ & $27.50_{\pm 3.0}$ & \textbf{--} & $39.03_{\pm 2.6}$ & $36.15_{\pm 2.7}$ \\
AdverIN~\cite{Zhang2026AdverIN} & $44.20_{\pm 2.3}$ & $29.01_{\pm 2.9}$ & \textbf{--} & $41.17_{\pm 2.4}$ & $38.13_{\pm 2.5}$ \\
ConDiSR~\cite{siddiquee2025condisr} & $44.43_{\pm 2.2}$ & $29.16_{\pm 2.8}$ & \textbf{--} & $41.37_{\pm 2.3}$ & $38.32_{\pm 2.4}$ \\
RAS$^4$DG~\cite{Qiao_2024_MICCAI} & $\underline{47.21}_{\pm 2.0}$ & $\underline{30.30}_{\pm 2.7}$ & \textbf{--} & $\underline{43.83}_{\pm 2.2}$ & $\underline{40.45}_{\pm 2.3}$ \\
\rowcolor{OurResult!15} \textbf{Ours} & $\mathbf{50.90}_{\pm \mathbf{1.7}}$ & $\mathbf{33.40}_{\pm \mathbf{2.4}}$ & \textbf{--} & $\mathbf{47.40}_{\pm \mathbf{1.9}}$ & $\mathbf{43.90}^{\dagger}_{\pm \mathbf{2.0}}$ \\
\midrule
\multicolumn{6}{c}{\cellcolor{gray!5}\textit{Source: Messidor-2}} \\

ERM & $40.73_{\pm 2.5}$ & $25.47_{\pm 3.0}$ & $40.66_{\pm 2.5}$ & \textbf{--} & $35.62_{\pm 2.7}$ \\
AdverIN~\cite{Zhang2026AdverIN} & $42.96_{\pm 2.3}$ & $26.87_{\pm 2.9}$ & $42.89_{\pm 2.4}$ & \textbf{--} & $37.57_{\pm 2.5}$ \\
ConDiSR~\cite{siddiquee2025condisr} & $43.18_{\pm 2.2}$ & $27.00_{\pm 2.9}$ & $43.11_{\pm 2.3}$ & \textbf{--} & $37.76_{\pm 2.5}$ \\
RAS$^4$DG~\cite{Qiao_2024_MICCAI} & $\underline{45.83}_{\pm 2.1}$ & $\underline{27.91}_{\pm 2.8}$ & $\underline{45.75}_{\pm 2.2}$ & \textbf{--} & $\underline{39.83}_{\pm 2.4}$ \\
\rowcolor{OurResult!15} \textbf{Ours} & $\mathbf{49.47}_{\pm \mathbf{1.8}}$ & $\mathbf{30.94}_{\pm \mathbf{2.5}}$ & $\mathbf{49.39}_{\pm \mathbf{1.9}}$ & \textbf{--} & $\mathbf{43.26}^{\dagger}_{\pm \mathbf{2.1}}$ \\
\bottomrule
\end{tabular}%
}
\end{minipage}
\end{table*}

\section{Results}
\label{sec:results}

Table~\ref{tab:full_comparison} reports the most stringent setting, where training is restricted to a \emph{single} source domain and evaluation spans all unseen domains. On Camelyon17-WILDS, PhaseAT delivers large, consistent gains over ERM and recent DG baselines, with the strongest improvements arising when the source center encourages the most pronounced center-specific shortcuts, for example, when trained on $C_2$, the average accuracy increases from $52.08$ (ERM) to $74.31$. Crucially, the gains are distributed across multiple target hospitals rather than concentrated on one, indicating reduced reliance on stain/texture-specific correlations and stronger sensitivity to morphology. On the Diabetic Retinopathy benchmark, PhaseAT similarly improves transfer across all source datasets (e.g., Aptos as source: $37.04 \rightarrow 44.99$ avg.), suggesting that enforcing stability to structured phase variations provides a robust inductive bias across heterogeneous acquisition conditions (camera/sensor, illumination, and preprocessing). Over the five shared seeds, gains marked $^{\dagger}$ are significant against the strongest baseline (paired $t$-test, $p<0.05$; e.g., SDG $C_0$: $\Delta{=}{+}2.24$, $95\%$ CI $[0.3,4.1]$).

\begin{table*}[t]
\centering
\caption{Performance comparison in the MSDG setting. Reported values are the accuracy(\%) mean $\pm$ standard deviation. Best and second-best per column are in \textbf{bold} and \underline{underline}; $^{\dagger}$ marks a significant gain over the strongest baseline (paired $t$-test, $p{<}0.05$).}
\label{tab:msdg_comparison}

\setlength{\tabcolsep}{2.0pt}

\begin{minipage}[t]{0.53\textwidth}
    \centering
    \subcaption{Camelyon17-WILDS}
    \resizebox{\textwidth}{!}{%
    \begin{tabular}{lcccccc}
        \toprule
        \rowcolor{gray!10} \textbf{\textsc{Method}} & $C_0$ & $C_1$ & $C_2$ & $C_3$ & $C_4$ & \textbf{\textsc{AVG.}} \\
        \midrule
        ERM & $78.89_{\pm 0.5}$ & $84.57_{\pm 1.1}$ & $67.24_{\pm 0.5}$ & $84.47_{\pm 0.2}$ & $74.08_{\pm 1.7}$ & $77.9$ \\
AdverIN~\cite{Zhang2026AdverIN} & $81.52_{\pm 0.6}$ & $87.78_{\pm 1.0}$ & $68.97_{\pm 0.7}$ & $88.17_{\pm 0.4}$ & $76.19_{\pm 1.6}$ & $80.5$ \\
ConDiSR~\cite{siddiquee2025condisr} & $81.67_{\pm 0.4}$ & $87.90_{\pm 0.7}$ & $\underline{72.12}_{\pm 0.2}$ & $86.79_{\pm 0.3}$ & $\mathbf{80.06}_{\pm \mathbf{0.3}}$ & $\underline{81.7}$ \\
RAS$^4$DG~\cite{Qiao_2024_MICCAI} & $\underline{83.62}_{\pm 0.5}$ & $\mathbf{89.10}_{\pm \mathbf{0.8}}$ & $69.31_{\pm 0.4}$ & $\underline{88.24}_{\pm 0.6}$ & $77.60_{\pm 0.5}$ & $81.6$ \\
\rowcolor{OurResult!15} \textbf{Ours} & $\mathbf{90.49}_{\pm \mathbf{0.5}}$ & $\underline{87.97}_{\pm 0.7}$ & $\mathbf{76.86}_{\pm \mathbf{0.3}}$ & $\mathbf{94.61}_{\pm \mathbf{0.5}}$ & $\underline{78.58}_{\pm 0.4}$ & $\mathbf{85.7}^{\dagger}$ \\
        \bottomrule
    \end{tabular}%
    }
\end{minipage}
\hfill
\begin{minipage}[t]{0.43\textwidth}
    \centering
    \subcaption{Diabetic Retinopathy}
    \resizebox{\textwidth}{!}{%
    \begin{tabular}{lccccc}
        \toprule
        \rowcolor{gray!10} \textbf{\textsc{Method}} & \textsc{Aptos} & \textsc{EyePACS} & \textsc{Messidor} & \textsc{Messidor2} & \textbf{\textsc{AVG.}} \\
        \midrule
        ERM & $55.7_{\pm 1.2}$ & $43.4_{\pm 0.8}$ & $54.3_{\pm 1.6}$ & $53.8_{\pm 0.4}$ & $51.8$ \\
AdverIN~\cite{Zhang2026AdverIN} & $58.6_{\pm 1.2}$ & $48.0_{\pm 0.9}$ & $58.3_{\pm 1.5}$ & $53.4_{\pm 0.5}$ & $54.6$ \\
ConDiSR~\cite{siddiquee2025condisr} & $59.4_{\pm 0.5}$ & $48.8_{\pm 0.4}$ & $59.2_{\pm 0.8}$ & $54.2_{\pm 0.2}$ & $55.4$ \\
RAS$^4$DG~\cite{Qiao_2024_MICCAI} & $\underline{60.9}_{\pm 0.9}$ & $\underline{50.2}_{\pm 0.6}$ & $\underline{60.6}_{\pm 1.1}$ & $\underline{55.6}_{\pm 0.3}$ & $\underline{56.8}$ \\
\rowcolor{OurResult!15} \textbf{Ours} & $\mathbf{62.4}_{\pm \mathbf{0.8}}$ & $\mathbf{51.1}_{\pm \mathbf{0.5}}$ & $\mathbf{62.1}_{\pm \mathbf{1.0}}$ & $\mathbf{56.8}_{\pm \mathbf{0.3}}$ & $\mathbf{58.1}^{\dagger}$ \\
        \bottomrule
    \end{tabular}%
    }
\end{minipage}

\end{table*}

Following the SDG results in Table~\ref{tab:full_comparison}, Table~\ref{tab:msdg_comparison} shows that the same phase-aware robustness mechanism remains effective when multiple source domains are available, albeit with a different gain regime. In SDG, improvements are often larger because single-source training amplifies center/dataset-specific shortcuts. In MSDG, pooling multiple sources already exposes the model to broader appearance variability, so the incremental gains are smaller but more stable. On Camelyon17-WILDS, PhaseAT achieves the best overall average ($85.7\%$), substantially improving over ERM ($77.9\%$) and surpassing recent DG methods clustered around $\sim 81$--$82\%$, with powerful benefits on shift-sensitive held-out hospitals (notably $C_2$), thereby improving worst-case behavior across centers. On Diabetic Retinopathy, PhaseAT likewise provides consistent improvements across all held-out datasets, increasing the average from $51.8\%$ to $58.1\%$ and outperforming the strongest baseline (RAS$^4$DG at $56.8\%$), indicating that enforcing stability to phase-driven structural perturbations complements multi-source supervision by discouraging reliance on dataset-specific color/contrast statistics and promoting geometry-aligned cues that transfer across acquisition pipelines. \\

\begin{figure*}[t!]
  \centering
  \includegraphics[width=0.90\linewidth]{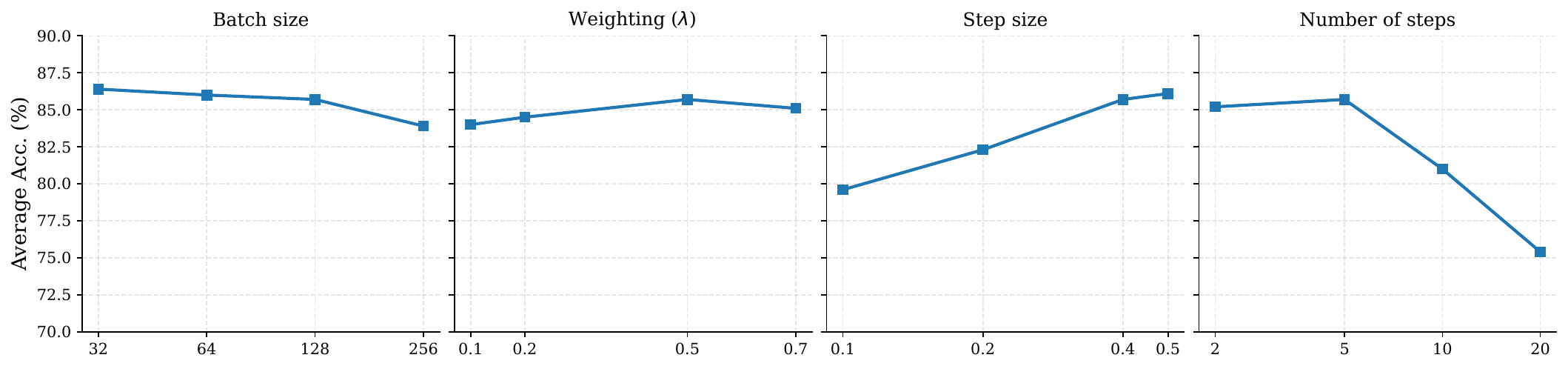}
  \caption{Ablation studies for the main hyperparameters on the average accuracy over the 5 centers in MSDG setting on Camelyon17-WILDS.}
  \label{fig:ablation}
\end{figure*}

\noindent\textbf{Ablation Study.} Figure \ref{fig:ablation} highlights both robustness and practicality. PhaseAT is comparatively stable as the batch size increases, consistent with the phase-perturbed supervision acting as an additional regularizer rather than relying on fragile optimization conditions. Performance is also insensitive to small variations in the loss weighting, with a broad optimum around $\lambda \approx 0.5$, indicating that the clean/phase-perturbed trade-off is not finely tuned. Moderate phase step sizes are beneficial, while overly many inner steps eventually degrade performance, consistent with perturbations becoming too strong and drifting away from the desired ``style-controlled'' stress test. Overall, the method exhibits a wide stable operating region and achieves its best results with moderate perturbation strength and a small number of steps.

\begin{table*}[t!]
\centering
\caption{Ablation Study on Camelyon17-WILDS in MSDG setting.}
\label{tab:ablation_split}

\begin{minipage}[t]{0.48\linewidth}
    \centering
    \subcaption{Backbone Performance}
    \resizebox{\linewidth}{!}{%
    \begin{tabular}{lcccccc}
        \toprule
        \rowcolor{gray!10}  & $C_0$ & $C_1$ & $C_2$ & $C_3$ & $C_4$ & Avg. \\
        \midrule
        ERM\_ResNet  & 78.83 & 83.56 & 79.00 & 90.66 & 60.39 & 78.49 \\
        \rowcolor{OurResult!15} Ours\_ResNet & 91.20 & 86.16 & 81.53 & 91.66 & 62.77 & 82.66 \\
        ERM\_ViT      & 74.20 & 78.40 & 62.10 & 86.50 & 45.30 & 69.30 \\
        \rowcolor{OurResult!15} Ours\_ViT     & 84.10 & 81.20 & 66.50 & 89.10 & 50.40 & 74.26 \\
        \midrule
        \multicolumn{7}{l}{\textbf{Densenet Imagenet Pretrained}} \\
        \midrule
        ERM    & 94.48 & 90.45 & 82.56 & 93.43 & 92.55 & 90.69 \\
        \rowcolor{OurResult!15} Ours   & 95.39 & 91.52 & 88.80 & 95.68 & 86.90 & 91.66 \\
        \bottomrule
    \end{tabular}%
    }
\end{minipage}
\hfill 
\begin{minipage}[t]{0.48\linewidth}
    \centering
    \subcaption{Component Ablation}
    \resizebox{\linewidth}{!}{%
    \begin{tabular}{lcccccc}
        \toprule
        \rowcolor{gray!10}  & $C_0$ & $C_1$ & $C_2$ & $C_3$ & $C_4$ & Avg. \\
        \midrule
        Amplitude-only   & 79.13 & 84.73 & 65.95 & 84.40 & 75.23 & 77.89 \\
        Phase+Amplitude & 80.30 & 85.46 & 66.12 & 84.52 & 76.12 & 78.50 \\
        No Mask     & 60.01 & 81.68 & 55.82 & 49.93 & 34.92 & 56.47 \\
        RGB         & 85.61 & 85.65 & 61.21 & 86.01 & 74.62 & 78.62 \\
        YCbCr-all   & 79.86 & 86.14 & 73.18 & 90.54 & 77.64 & 81.47 \\
        Random  Phase   & 89.57 & 87.87 & 64.01 & 91.82 & 76.22 & 81.90 \\
        \bottomrule
    \end{tabular}%
    }
\end{minipage}
\end{table*}

Table \ref{tab:ablation_split}(a) demonstrates architectural generality: PhaseAT improves ResNet-18 ($78.49 \rightarrow 82.66$ avg.) and ViT-Base ($69.30 \rightarrow 74.26$ avg.), indicating that the benefit is not tied to convolutional texture priors and persists for transformer backbones. With ImageNet-pretrained DenseNet, gains are smaller ($90.69 \rightarrow 91.66$), suggesting that generic pretraining already contributes partial robustness, while PhaseAT still adds complementary invariances. 

Table \ref{tab:ablation_split} (b) clarifies which choices matter. PhaseAT perturbs \emph{phase} while \emph{fixing} amplitude; when we instead (i) fix phase and update amplitude only or (ii) update both phase and amplitude (no amplitude fixing), performance drops, consistent with the intended ``structure under matched style'' shift. The saliency mask is particularly important: without it, performance degrades sharply, indicating that concentrating updates on phase-sensitive frequencies stabilizes training. Operating on luminance is consistently advantageous: RGB-space perturbations degrade performance, and perturbing all YCbCr channels recovers only partially, reinforcing the choice to emphasize geometry and avoid unnecessary chromatic distortions. The ``Random Phase'' variant (sampling a bounded random phase offset within the same $\ell_\infty$ budget, instead of gradient-based inner maximization) remains strong, showing that phase jitter is a useful regularizer, but the gradient-guided procedure provides the best overall generalization.

\section{Conclusion and Future Work}
We introduce \textsc{PhaseAT}, a phase-aware adversarial training framework that perturbs the Fourier phase of the luminance channel while fixing amplitude, generating structure-stressing views with matched appearance and reducing reliance on domain-specific texture shortcuts. \textsc{PhaseAT} improves transfer on Camelyon17-WILDS and diabetic retinopathy, achieving up to +22\% accuracy gains over ERM in single-source settings and +9\% multi-source.
Our evaluation is currently limited to 2D classification and color medical images (YCbCr); investigating performance on grayscale modalities remains an important limitation. Future work will extend phase-constrained adversarial views to dense prediction, 3D modalities, and broader acquisition shifts, and study more adaptive frequency masking/selection, combinations with complementary amplitude-/style-based DG objectives, and efficiency improvements that reduce training overhead while preserving robustness.

\subsubsection{\discintname}
The authors have no competing interests to declare that are
relevant to the content of this article.

%

\bibliographystyle{splncs04}
\bibliography{refs}

@article{Zhang2026AdverIN,
  title={AdverIN: Monotonic adversarial intensity attack for domain generalization in medical image segmentation},
  author={Zhang, Zheyuan and Wang, Bin and Yao, Lanhong and Keles, Elif and Jha, Debesh and Antalek, Matthew and Durak, Gorkem and Medetalibeyoglu, Alpay and Spampinato, Concetto and Turkbey, Baris and Gong, Boqing and Bagci, Ulas},
  journal={Medical Image Analysis},
  volume={107},
  pages={103848},
  year={2026},
  doi={10.1016/j.media.2025.103848}
}

@inproceedings{Chen2022MaxStyle,
  title={MaxStyle: Adversarial Style Composition for Robust Medical Image Segmentation},
  author={Chen, Chen and Li, Zeju and Ouyang, Cheng and Sinclair, Matthew and Bai, Wenjia and Rueckert, Daniel},
  booktitle={Medical Image Computing and Computer Assisted Intervention -- MICCAI 2022},
  series={Lecture Notes in Computer Science},
  volume={13435},
  pages={151--161},
  year={2022},
  doi={10.1007/978-3-031-16443-9_15}
}

@inproceedings{Safdari2025MixStyleFlow,
  title={{MixStyleFlow: Domain Generalization in Medical Image Segmentation using Normalizing Flows}},
  author={Safdari, Reza and Nikouei Mahani, Mohammad-Ali and Koohi-Moghadam, Mohamad and Bae, Kyongtae Tyler},
  booktitle={Medical Image Computing and Computer Assisted Intervention -- MICCAI 2025},
  series={Lecture Notes in Computer Science},
  volume={15962},
  pages={376--385},
  year={2025},
  doi={10.1007/978-3-032-04947-6_36}
}

@inproceedings{CheXi_ConStyX_MICCAI2025,
  author = {Chen, Xi and Shen, Zhiqiang and Cao, Peng and Yang, Jinzhu and Zaiane, Osmar R.},
  title = {ConStyX: Content Style Augmentation for Generalizable Medical Image Segmentation},
  booktitle = {Medical Image Computing and Computer Assisted Intervention -- MICCAI 2025},
  series = {Lecture Notes in Computer Science},
  volume = {15964},
  pages = {100--110},
  year = {2025},
  doi = {10.1007/978-3-032-04971-1_10}
}

@inproceedings{Qiao_2024_MICCAI,
  author = {Qiao, Qiang and Wang, Wenyu and Qu, Meixia and Su, Kun and Jiang, Bin and Guo, Qiang},
  title = {Medical Image Segmentation via Single-Source Domain Generalization with Random Amplitude Spectrum Synthesis},
  booktitle = {Medical Image Computing and Computer Assisted Intervention -- MICCAI 2024},
  series = {Lecture Notes in Computer Science},
  volume = {15009},
  pages = {435--445},
  year = {2024},
  publisher = {Springer},
  doi = {10.1007/978-3-031-72114-4_42}
}

@inproceedings{Yi_2024_MICCAI,
  author = {Yi, Jingjun and Bi, Qi and Zheng, Hao and Zhan, Haolan and Ji, Wei and Huang, Yawen and Li, Shaoxin and Li, Yuexiang and Zheng, Yefeng and Huang, Feiyue},
  title = {Hallucinated Style Distillation for Single Domain Generalization in Medical Image Segmentation},
  booktitle = {Medical Image Computing and Computer Assisted Intervention -- MICCAI 2024},
  series = {Lecture Notes in Computer Science},
  volume = {15010},
  pages = {438--448},
  year = {2024},
  publisher = {Springer},
  doi = {10.1007/978-3-031-72117-5_41}
}

@InProceedings{Jiang2024EdgeDG,
  title = 	 {Train Once, Deploy Anywhere: Edge-Guided Single-source Domain Generalization for Medical Image Segmentation},
  author =       {Jiang, Jun and Gu, Shi},
  booktitle = 	 {Proceedings of The 7nd International Conference on Medical Imaging with Deep Learning},
  pages = 	 {722--741},
  year = 	 {2024},
  editor = 	 {Burgos, Ninon and Petitjean, Caroline and Vakalopoulou, Maria and Christodoulidis, Stergios and Coupe, Pierrick and Delingette, Hervé and Lartizien, Carole and Mateus, Diana},
  volume = 	 {250},
  series = 	 {Proceedings of Machine Learning Research},
  month = 	 {03--05 Jul},
  publisher =    {PMLR},
  url = 	 {https://proceedings.mlr.press/v250/jiang24a.html},
}

@inproceedings{Hu2024MAP,
  title={MAP: Domain Generalization via Meta-Learning on Anatomy-Consistent Pseudo-Modalities},
  author={Hu, Dewei and Li, Hao and Liu, Han and Yao, Xing and Wang, Jiacheng and Oguz, Ipek},
  booktitle={Medical Image Computing and Computer Assisted Intervention -- MICCAI 2023 Workshops (MedAGI)},
  series={Lecture Notes in Computer Science},
  volume={14393},
  year={2023},
  publisher={Springer},
  doi={10.1007/978-3-031-47401-9_18}
}

@inproceedings{Su2023SSDA,
  title={Rethinking Data Augmentation for Single-Source Domain Generalization in Medical Image Segmentation},
  author={Su, Zixian and Yao, Kai and Yang, Xi and Wang, Qiufeng and Sun, Jie and Huang, Kaizhu},
  booktitle={Proceedings of the AAAI Conference on Artificial Intelligence},
  volume={37},
  number={2},
  pages={2366--2374},
  year={2023},
  doi={10.1609/aaai.v37i2.25332}
}

@inproceedings{Lee2023DACNorm,
  title={Decompose, Adjust, Compose: Effective Normalization by Playing With Frequency for Domain Generalization},
  author={Lee, Sangrok and Bae, Jongseong and Kim, Ha Young},
  booktitle={Proceedings of the IEEE/CVF Conference on Computer Vision and Pattern Recognition (CVPR)},
  pages={11776--11785},
  year={2023},
  doi={10.1109/CVPR52729.2023.01134}
}

@inproceedings{Zhang_2025_ICCV,
  author = {Zhang, Zuyu and Chen, Ning and Liu, Yongshan and Zhang, Qinghua and Zhang, Xu},
  title = {Adversarial Data Augmentation for Single Domain Generalization via Lyapunov Exponent-Guided Optimization},
  booktitle = {Proceedings of the IEEE/CVF International Conference on Computer Vision (ICCV)},
  year = {2025},
  url = {https://arxiv.org/abs/2507.04302}
}

@inproceedings{Teterwak_2025_ICLR,
  author = {Teterwak, Piotr and Saito, Kuniaki and Tsiligkaridis, Theodoros and Plummer, Bryan A. and Saenko, Kate},
  title = {Is Large-Scale Pretraining the Secret to Good Domain Generalization?},
  booktitle = {Proceedings of the International Conference on Learning Representations (ICLR)},
  year = {2025},
  url = {https://openreview.net/forum?id=wCOJpXm0Me}
}

@inproceedings{zhang2024fourier,
  author    = {Zhang, Liangqi and Luo, Yihao and Shen, Haibo and Wang, Tianjiang},
  title     = {A Fourier Perspective of Feature Extraction and Adversarial Robustness},
  booktitle = {Proceedings of the Thirty-Third International Joint Conference on Artificial Intelligence (IJCAI)},
  pages     = {1715--1723},
  year      = {2024},
  doi       = {10.24963/ijcai.2024/190}
}

@inproceedings{zhou2023phaseaware,
  author    = {Zhou, Dawei and Wang, Nannan and Yang, Heng and Gao, Xinbo and Liu, Tongliang},
  title     = {Phase-aware Adversarial Defense for Improving Adversarial Robustness},
  booktitle = {Proceedings of the 40th International Conference on Machine Learning (ICML)},
  series    = {Proceedings of Machine Learning Research},
  volume    = {202},
  pages     = {42724--42741},
  year      = {2023}
}

@inproceedings{geirhos2019imagenettexture,
  title     = {ImageNet-trained CNNs are biased towards texture; increasing shape bias improves accuracy and robustness},
  author    = {Geirhos, Robert and Rubisch, Patricia and Michaelis, Claudio and Bethge, Matthias and Wichmann, Felix A. and Brendel, Wieland},
  booktitle = {International Conference on Learning Representations (ICLR)},
  year      = {2019},
  note      = {ICLR 2019 (oral)},
  url       = {https://arxiv.org/abs/1811.12231}
}

@misc{wirtinger,
  title={A short tutorial on Wirtinger Calculus with applications in quantum information},
  author={Kelvin Koor and Yixian Qiu and Leong Chuan Kwek and Patrick Rebentrost},
  year={2023},
  eprint={2312.04858},
  archivePrefix={arXiv},
  primaryClass={quant-ph},
  url={https://arxiv.org/abs/2312.04858}
}

@article{bandi2019camelyon17,
  title   = {From Detection of Individual Metastases to Classification of Lymph Node Status at the Patient Level: The {CAMELYON17} Challenge},
  author  = {B{\'a}ndi, P{\'e}ter and Geessink, Oscar and Manson, Quirine and Van Dijk, Marcory and Balkenhol, Maschenka and Hermsen, Meyke and Bejnordi, Babak Ehteshami and Lee, Byungjae and Paeng, Kyunghyun and Zhong, Aoxiao and Li, Quanzheng and Zanjani, Farhad Ghazvinian and Zinger, Svitlana and others},
  journal = {IEEE Transactions on Medical Imaging},
  volume  = {38},
  number  = {2},
  pages   = {550--560},
  year    = {2019},
  doi     = {10.1109/TMI.2018.2867350}
}

@misc{aptos2019blindness,
  title        = {{APTOS} 2019 Blindness Detection},
  howpublished = {Kaggle Competition},
  year         = {2019},
  url          = {https://www.kaggle.com/c/aptos2019-blindness-detection},
  note         = {Accessed: 2026-01-27}
}

@misc{kaggle2015dr,
  title        = {Diabetic Retinopathy Detection},
  howpublished = {Kaggle Competition},
  year         = {2015},
  url          = {https://www.kaggle.com/c/diabetic-retinopathy-detection},
  note         = {Accessed: 2026-01-27}
}

@article{decenciere2014messidor,
  title   = {Feedback on a Publicly Distributed Image Database: The {MESSIDOR} Database},
  author  = {Decenci{\`e}re, Etienne and Zhang, Xiwei and Cazuguel, Guy and Lay, Bruno and Cochener, B{\'e}atrice and Trone, Caroline and Gain, Philippe and Ordonez, Richard and Massin, Pascale and Erginay, Ali and Charton, B{\'e}atrice and Klein, Jean-Claude},
  journal = {Image Analysis \& Stereology},
  volume  = {33},
  number  = {3},
  pages   = {231--234},
  year    = {2014},
  doi     = {10.5566/ias.1155}
}

@misc{messidor2,
  title        = {{MESSIDOR-2} Dataset},
  howpublished = {ADCIS (dataset access page)},
  url          = {https://www.adcis.net/en/third-party/messidor2/},
  note         = {Accessed: 2026-01-27}
}

@misc{gulrajani2020domainbed,
  title         = {In Search of Lost Domain Generalization},
  author        = {Gulrajani, Ishaan and Lopez-Paz, David},
  year          = {2020},
  eprint        = {2007.01434},
  archivePrefix = {arXiv},
  primaryClass  = {cs.LG},
  url           = {https://arxiv.org/abs/2007.01434}
}

@inproceedings{siddiquee2025condisr,
  title={ConDiSR: Contrastive Disentanglement and Style Regularization for Single Domain Generalization},
  author={Matsun, Aleksandr and Saeed, Numan and Maani, Fadillah Adamsyah and Yaqub, Mohammad},
  booktitle={Proceedings of the IEEE/CVF Winter Conference on Applications of Computer Vision (WACV)},
  pages={2881--2889},
  year={2025},
  publisher={IEEE}
}

@INPROCEEDINGS{pestana2020ychannel,
  author={Pestana, Camilo and Akhtar, Naveed and Liu, Wei and Glance, David and Mian, Ajmal},
  booktitle={2021 International Joint Conference on Neural Networks (IJCNN)}, 
  title={Adversarial Attacks and Defense on Deep Learning Classification Models using YCbCr Color Images}, 
  year={2021},
  volume={},
  number={},
  pages={1-9},
  doi={10.1109/IJCNN52387.2021.9533495}}
\end{document}